\documentclass{article}
\usepackage{CJKutf8}
\usepackage[utf8]{inputenc}
\usepackage{authblk}
\usepackage{setspace}
\usepackage[margin=1.25in]{geometry}
\usepackage{graphicx}
\graphicspath{ {./figures/} }
\usepackage{subcaption}
\usepackage{amsmath}
\usepackage{lineno}
\usepackage{hyperref}
\usepackage{cleveref} 
\usepackage{tabularx}
\usepackage[figuresright]{rotating}

\usepackage[style=nejm, 
citestyle=numeric-comp,
sorting=none]{biblatex}
\title{
Agent4S: The Transformation of Research Paradigms from the Perspective of Large Language Models}

\author[1,2*$\dag$]{Boyuan Zheng}
\author[3,4$\dag$]{Zerui Fang}
\author[1,5$\dag$]{Zhe Xu}

\author[3,1,2]{Rui Wang}
\author[1,2]{Yiwen Chen}
\author[6,7]{Cunshi Wang}
\author[8,9]{Mengwei Qu}
\author[10]{Lei Lei}
\author[11,12]{Zhen Feng}
\author[13]{Yan Liu}

\author[6,14*]{Yuyang Li}
\author[15*]{Mingzhou Tan}
\author[16*]{Jiaji Wu}
\author[17,12*]{Jianwei Shuai}
\author[4*]{Jia Li}
\author[1,18,12*]{Fangfu Ye}

\affil[1]{\small Beijing National Laboratory for Condensed Matter Physics, Institute of Physics, Chinese Academy of Sciences, Beijing, China}
\affil[2]{Beijing Gongyu Zhiyan Technology Co., Ltd, Beijing, China}
\affil[3]{School of Advanced Interdisciplinary Sciences, University of Chinese Academy of Science, Beijing, China}
\affil[4]{Smart Sensing Chip and System R\&D Center, Institute of Microelectronics of the Chinese Academy of Sciences, Beijing, China}
\affil[5]{Key Laboratory of Material Physics, Ministry of Education, School of Physics and Microelectronics, Zhengzhou University, Zhengzhou, China}
\affil[6]{Key Laboratory of Optical Astronomy, National Astronomical Observatories, Chinese Academy of Sciences, Beijing, China}
\affil[7]{College of Astronomy and Space Sciences, University of Chinese Academy of Sciences, Beijing, China}
\affil[8]{State Key Laboratory of Isotope Geochemistry, Guangzhou Institute of Geochemistry, Chinese Academy of Sciences, Guangzhou, Guangdong China}
\affil[9]{College of Earth and Planetary Sciences, University of Chinese Academy of Sciences, Beijing, China}
\affil[10]{Alibaba Cloud, Hangzhou, Zhejiang, China}
\affil[11]{College of Information and Engineering, Wenzhou Medical University, Wenzhou, Zhejiang, China}
\affil[12]{Wenzhou Institute, University of Chinese Academy of Sciences, Wenzhou, Zhejiang, China}
\affil[13]{Department of science and development, Chinese academy of sciences, Beijing, China}
\affil[14]{College of Astronomy and Space Sciences, University of Chinese Academy of Sciences, Beijing, China}
\affil[15]{Emotion Machine (Beijing) Technology Co., Ltd., Beijing, China}
\affil[16]{School of Electronic Engineering, Xidian University, Xi’an, Shaanxi, China}
\affil[17]{Oujiang Laboratory (Zhejiang Lab for Regenerative Medicine, Vision and Brain Health), Wenzhou, Zhejiang, China}
\affil[18]{School of Physical Sciences, University of Chinese Academy of Sciences, Beijing, China}

\affil[*]{Corresponding authors are as follows: Fangfu Ye(fye@iphy.ac.cn), Jia Li(lijia@ime.ac.cn), Yuyang Li(liyuyang22@mails.ucas.ac.cn), Jianwei Shuai(jianweishuai@xmu.edu.cn), Mingzhou Tan(SwanLab.cn), Jiaji Wu(wujj@mail.xidian.edu.cn), Boyuan Zheng(zhengboyuan17@mails.ucas.ac.cn)}
\affil[$\dag$]{Contributed equally (Co-first authorship)}

\date{}

\begin{document}

\maketitle
\quad

\quad

\begin{abstract}
While AI for Science (AI4S) serves as an analytical tool in the current research paradigm, it doesn't solve its core inefficiency. We propose "Agent for Science" (Agent4S)—the use of LLM-driven agents to automate the entire research workflow—as the true Fifth Scientific Paradigm. This paper introduces a five-level classification for Agent4S, outlining a clear roadmap from simple task automation to fully autonomous, collaborative "AI Scientists." This framework defines the next revolutionary step in scientific discovery.
\end{abstract}

\section{The Basic Characteristics and Development History of Scientific Research}
The essence of scientific research is to establish a set of methods to discover laws from data (phenomena), and then use these laws to guide production and transform the world. 

The four scientific paradigm revolutions\cite{hey2009the} shown in \Cref{table:1} are essentially transformations in methods of data acquisition and data processing.

The first scientific paradigm is the empirical paradigm, where the research paradigm of this stage is characterized by observation and induction. Data acquisition mainly relied on naked-eye observation and measurements with simple tools, such as the experimental methodology systematically expounded by Bacon in Novum Organum (1620)\cite{Bacon_Novum1620} and Galileo's telescopic observations\cite{taton2003planetary}. Data processing involved manual recording and simple statistics, establishing causal relationships through repeated observations and emphasizing the reproducibility and intuitiveness of experience.

The second scientific paradigm is the theoretical paradigm. With the development of precision instruments and mathematical tools, scientific research achieved a leap from qualitative to quantitative analysis. Data acquisition shifted from simple observation to controlled experiments and precise measurements, such as Newton’s prism experiments on the dispersion of light spectra in 1665–1666\cite{doi:10.1098/rstl.1671.0072} and Cavendish’s torsion balance experiment to measure the gravitational constant in 1797–1798\cite{doi:10.1098/rstl.1798.0022}. Mathematical formulas and equation solving were introduced into data processing, enabling the derivation of universal laws from limited experimental data through mathematical abstraction and establishing a "mathematical" scientific language system.

\begin{table}[bht]
\caption{Revolution of the four scientific paradigms}
\centering
\begin{tabularx}{\textwidth}{lXXX}
\hline
Paradigm & Key Characteristics & Data Acquisition Methods & Data Processing Methods \\ \hline
Empirical Paradigm & Observation-Induction & Naked-Eye Observation/Simple Instruments & Oral/Textual Records + Naive Induction \\
Theoretical Paradigm & Experimental-Quantitative Theory & Controlled Experiments and Precision Instruments & Mathematical Abstraction + Equation Solving \\
Computational Paradigm & Numerical Simulation & Computer-Generated Data & Numerical + High-Performance Computing \\
Data-Driven Paradigm & Big Data & Massive High-Dimensional Data from Sensors/Automated Devices & Machine Learning/Deep Learning \\
\hline
\end{tabularx}
\label{table:1}
\end{table}

The third scientific paradigm is the computational science paradigm. With the advent of electronic computers, scientific research broke through the physical limitations of traditional experiments. Data acquisition shifted from physical observation to computer-generated data, generating massive simulated data through mathematical models—examples include nuclear weapons test simulations, weather forecasting models, and molecular dynamics simulations. Data processing underwent a revolutionary shift from manual calculations to numerical computations: the introduction of high-performance computing (HPC) techniques made simulation research on complex systems feasible.

The fourth scientific paradigm is the data-intensive scientific paradigm\cite{gray2007escience}. With the development of sensor networks, the internet, and artificial intelligence technologies, scientific research has entered an era of "big data-driven" inquiry. Data acquisition now enables massive, high-dimensional, real-time collection through sensors, automated devices, and internet platforms—examples include gene sequencing, radio telescope arrays, and social network data streams. Data processing has evolved from statistical learning to machine learning and now deep learning, endowing computers with the ability to directly extract patterns from high-dimensional data spaces and achieving a fundamental shift from hypothesis-driven to data-driven research.

Each transformation has brought about corresponding scientific research processes and gradually improved the systematic methodologies of scientific research. The current scientific research process primarily involves: formulation of scientific questions, definition of research content, design of research plans, experimentation, and data processing with iteration.

With the continuous improvement of productivity, the rate of new data generation and the volume of knowledge derived from existing data have been growing exponentially. Under the framework of the fourth scientific research paradigm, a past contradiction was the mismatch between the increasing data dimensionality and insufficient data analysis methods—the curse of dimensionality\cite{hammer1962adaptive}. The development of deep learning has alleviated this contradiction to some extent\cite{bach2017breaking}. However, more fundamentally, from the perspective of productivity, a deeper contradiction lies in the growing volume of scientific research information and the inefficiency of existing research paradigms.

\section{Understanding AI4S from the Perspective of Scientific Data Processing}

\subsection{Definition of AI4S}

For the first contradiction, namely the mismatch between increasing data dimensionality and insufficient data analysis methods, AI algorithms—especially deep neural networks—as computational methods in the big data era that can handle higher-dimensional data compared to statistical learning and classical machine learning (or "high-order function fitting methods"), are capable of processing higher-dimensional data and uncovering more patterns within them. Their application in scientific research constitutes the classical definition of AI4S\cite{wang2023scientific}. For example, AlphaFold leverages the Evoformer algorithm to extract patterns from massive protein data on how amino acid interactions affect their distances, while deep learning potentials (DPMD) use neural networks to learn the relationships between molecular structures and potential energies.

\subsection{Development History and Current Status of AI4S}

Since expert systems first emerged in the 1980s and deep learning breakthroughs in image/speech domains around 2010 were rapidly applied to scientific computing, AI4S has undergone three critical leaps:
\begin{enumerate}
    \item In the symbolism stage, relying on expert-defined rules, it faced limitations in modeling complex systems;
    \item The statistical machine learning stage focused on low-dimensional feature engineering (e.g., support vector machines, random forests)—while enabling automatic modeling, it struggled with exponentially increasing data dimensions;
    \item The deep representation learning stage introduced "high-order function fitters" like convolutional networks, attention mechanisms, and graph neural networks into research scenarios, achieving breakthroughs in protein folding (AlphaFold 2/Evoformer), molecular potential energy surfaces (DPMD), inverse materials design (GraphGPT-Materials), etc.
\end{enumerate}

\section{Agent4S}

\subsection{Definition of Agent4S}

\begin{sidewaystable}[]
\caption{Agent4S Hierarchy}
\centering
\begin{tabularx}{\textwidth}{lXXXXXXXX}
\hline
Level & Definition & Agent Hierarchy & Underlying Agent Technology & Agent Technology Status & Research Phase & Paradigm Transformation Mode & Current Research Applications & Implementation Challenges \\\hline
L1 & Automation of a Single Tool & AI Agents (AI Proxies) & Prompt Engineering + FC + Workflow & Mature & Scientific Data Generation/Acquisition & Automated Tools for Fixed Processes & Preliminary Cases with Clear Development Trajectory & Hardware Digitization \\
L2 & Automation of Complex Processes &  &  &  &  &  &  & Data Transmission Robustness \\
L3 & Intelligence of a Single Process & Agentic AI (Intelligent Agent AI) & Reasoning + Context Engineering + MCP & Early Development & Scientific Data Analysis/Processing & Intelligent Collaborative Partner for Innovative Processes & No Clear Pattern Established & MCP-Enabled Research Tools and Context Engineering \\
L4 & Full-Process Intelligence (Single Laboratory) &  &  &  &  &  &  & Laboratory Hardware/Software Integration \\
L5 & Collaboration of Multiple Intelligent Processes (Multiple Laboratories) & Multi-Agent System & A2A (Agent-to-Agent) & Early Development & Interdisciplinary Research & Intelligent Network for Cross-Disciplinary Collaboration & No Clear Pattern Established & Breaking Disciplinary (Laboratory) Barriers \\ \hline
\end{tabularx}
\label{table:2}
\end{sidewaystable}

Starting from the development of Agent technology and combined with the existing paradigms of scientific research, we propose five levels of Agent4S, as shown in \Cref{table:2}.

The earliest form of agents originated from the LLM + Prompt mechanism, supplemented by Function Calling (FC) to map language outputs to API calls. This has been widely deployed in industry for customer service Q\&A and content generation; in scientific research, it manifests as the intelligent encapsulation of single scientific tools, such as literature retrieval, database query, image annotation, etc. (specific cases and references needed, note: the implementation must be a single agent). We define this as Level 1 (L1) of Agent4S—Automation of a Single Scientific Tool—referring to the automation of a simple task in research.

With the maturation of Task-/Workflow Orchestration frameworks (Airflow, Dagster, Ray Serve, etc.), LLM Agents can now maintain state and dependency relationships across multi-step tasks. Integrated "instruction-execution-callback" loops have been widely deployed in industrial scenarios such as robotic process automation (RPA), A/B testing, and advertising. The most typical scientific research practice is the "end-to-end data pipeline"—for example, integrating four steps of sequencing quality control, alignment, quantification, and statistics into a single-trigger automated task flow; or chaining structure generation, first-principles calculations, and database storage into one-click scripts for high-throughput materials computation (specific cases and references needed, note: implementation must be workflow-driven multi-agent, where workflows can be linear, parallel, or autonomously planned). We define this as Level 2 (L2) of Agent4S—Automation of Complex Scientific Pipelines—referring to orchestrating multiple L1 tools via workflow methods into reusable, low-human-intervention research pipelines to automate a complex fixed process in scientific research.

\begin{figure*}[tbp]
    \flushright
    \includegraphics[scale=0.045]{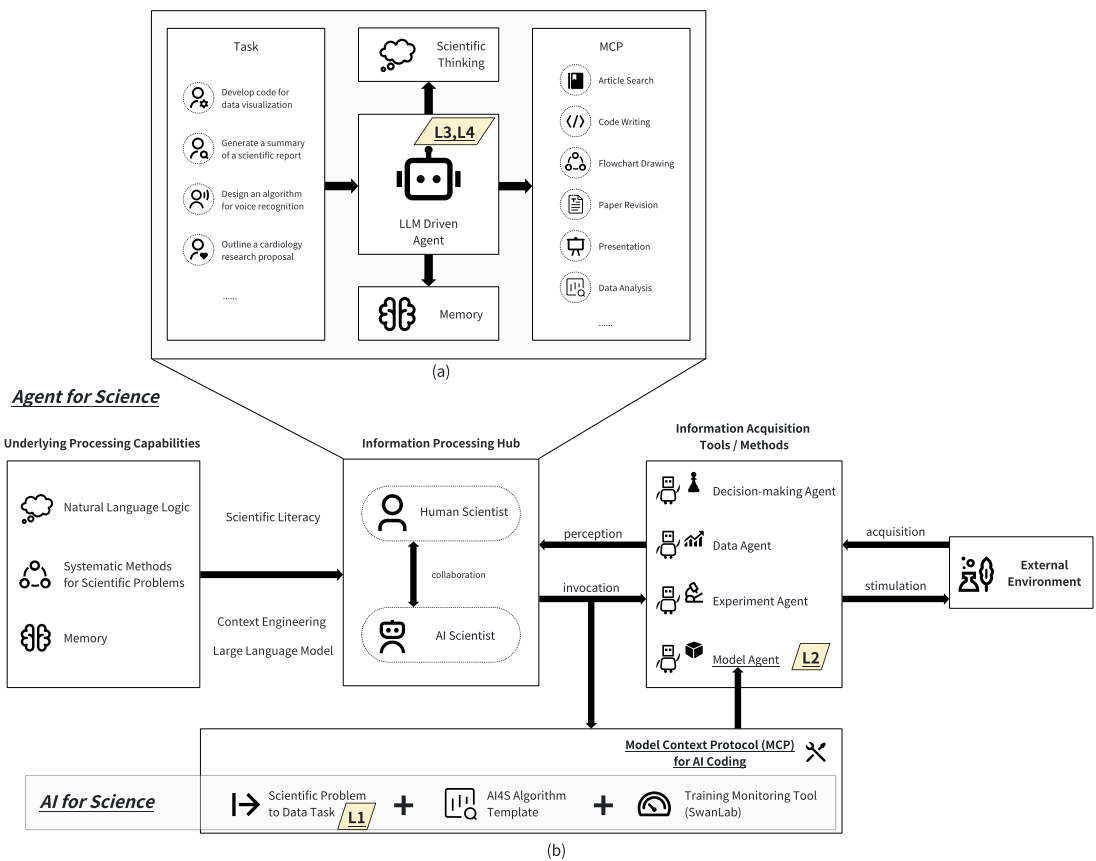}
    \caption{Technical framework}
    \label{fig:tech}
\end{figure*}

Currently, with the development of Reasoning Frameworks (ReAct, Tree-of-Thought, Graph-of-Thought) and Long-Context Memory Engineering, LLM Agents have evolved from "pipeline executors" to master agents capable of chain reasoning, real-time decision-making, and self-reflection. They achieve intelligent tool invocation through the Model-Context-Protocol (MCP), marking the transition from AI Agents to Agentic AI \cite{sapkota2025ai}. This paradigm has validated the feasibility of "observation-planning-iteration" closed-loop decision-making in industrial scenarios like AIOps, real-time risk control, and AI coding. Although AI-related technologies are newly developed and have limited applications in scientific research, it is foreseeable that the next stage will involve an AI Scientist with autonomous planning and long-term memory capabilities—capable of invoking various L2 tools via MCP to realize a closed-loop of "planning-tool usage-data analysis" for specific research processes, enabling intelligent iteration. We define this as Level 3 (L3) of Agent4S—Intelligent Single-Flow Research.

Looking ahead, as Agent capabilities in memory length, multi-step planning, and MCP invocation continue to advance—alongside the development of embodied intelligence and further integration/data connectivity across all hardware-software components in laboratory workflows—the L3-level AI Scientist will be able to participate in the full "hypothesis-experiment-analysis" closed loop of research projects and enable intelligent iteration. We define this as Level 4 (L4) of Agent4S—Lab-Scale Closed-Loop Autonomy.

Ultimately, with the advancement of full-process intelligence in each laboratory and relying on the A2A protocol, multiple super-agents will be able to communicate with each other, enabling the formation of an intelligent network for cross-disciplinary collaboration driven by multi-agent systems in scientific research. This constitutes the Level 5 (L5) stage of Agent4S and represents the final form of the fifth scientific paradigm driven by artificial intelligence.

From the perspective of general research processes, AI4S represents a more advanced method within the data analysis phase;
L1 of Agent4S refers to the automation of a simple tool used in a specific research sub-task (e.g., literature retrieval or data annotation), involving single-agent operations without cross-component integration.
L2 involves the automation of an entire research sub-process (e.g., a data pipeline combining sequencing QC and statistical analysis), differing from L1 in featuring a multi-agent-driven complete data closed loop orchestrated via workflow frameworks.
L3 denotes the intelligence of a specific research workflow, encompassing autonomous planning, tool invocation, data processing, and result collation. Its key distinction from L2 lies in the presence of a super-agent with context engineering capabilities that autonomously plans, orchestrates, and iterates the workflow.
L4 represents full-process intelligence in scientific research, covering the entire lifecycle from scientific question formulation, research design, hypothesis generation, experimental simulation, to data interpretation. It differs from L3 in enabling end-to-end participation across the entire research process rather than individual workflows.

In terms of future human-computer interaction models, as illustrated in \Cref{fig:tech}:
Levels L1 and L2 involve the automation of fixed processes, primarily applied to the historically machine-driven data generation/acquisition stages. Fundamentally, they remain tools within the research workflow.
Levels L3 and L4 represent the intelligence of innovative processes with stochastic and emergent characteristics, focusing on data analysis and interpretation. These constitute super-agents analogous to scientists—i.e., AI Scientists—capable of autonomous reasoning. A plausible future interaction model is scientists collaborating with AI Scientists as partners, jointly invoking various tools (L3/L4), which themselves may integrate numerous agents (L1/L2).
Level L5 refers to interactions among multiple AI Scientists, forming an intelligent network for cross-disciplinary research.

\begin{figure*}[bth]
    \centering
    \includegraphics[scale=0.22]{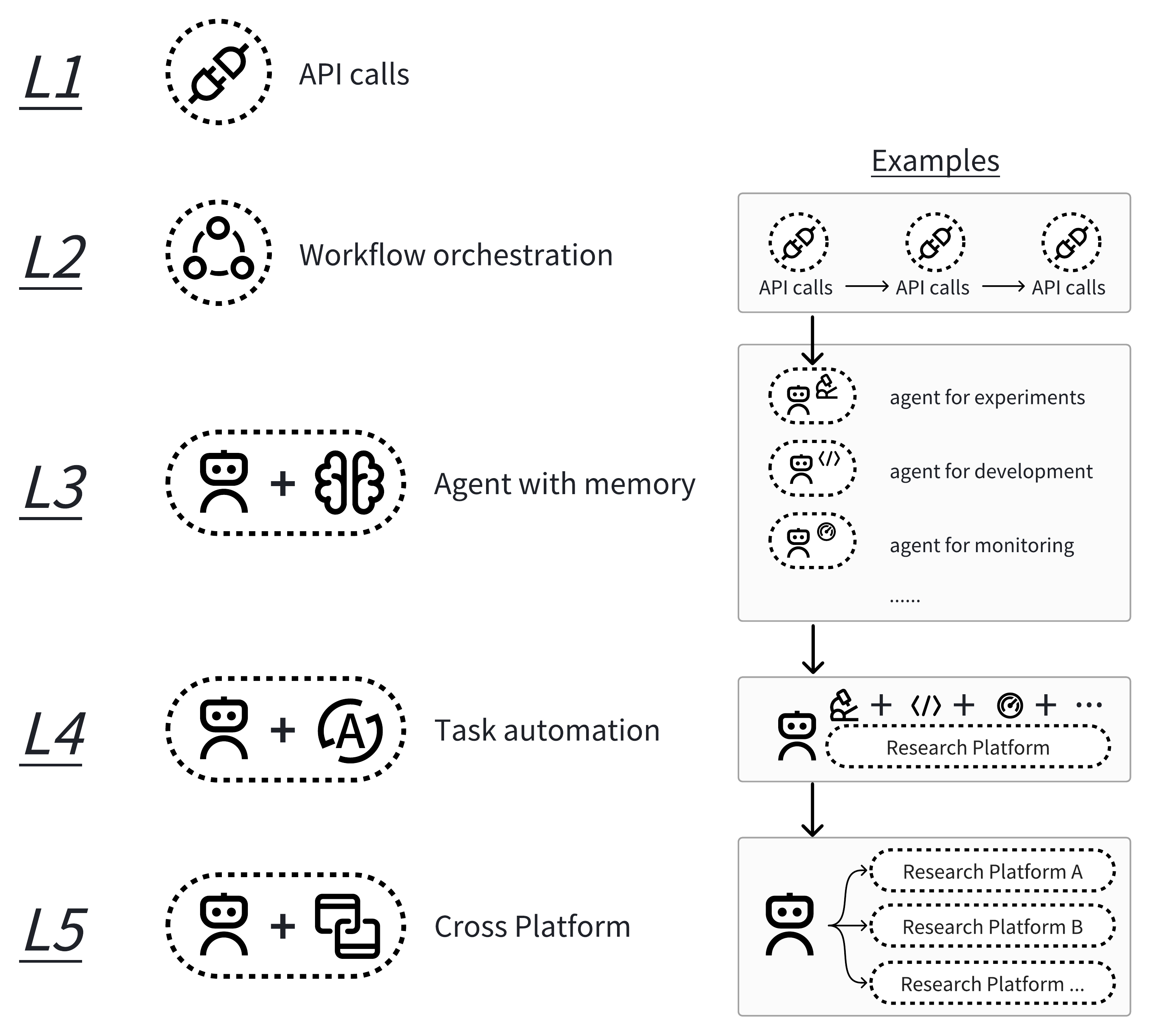}
    \caption{Five levels of Agent4S}
    \label{fig:l1tol5}
\end{figure*}

In terms of research organizational forms, as shown in \Cref{fig:l1tol5}:
L1 and L2 represent the automation of tools used by scientists in specific research sub-tasks, where agents remain as passive instruments within the workflow.
L3 marks a shift: agents evolve from "tools" to collaborative assistants with defined roles, capable of participating in targeted workflow stages under human guidance.
L4 elevates agents to the role of central coordinators in full experimental cycles—analogous to an AI project leader/laboratory director that oversees the entire research lifecycle, from proposal development to completion, with full situational awareness and autonomous decision-making capabilities.
L5 enables cross-disciplinary collaboration across multiple laboratories, where information flows through a network of super-agents, facilitating seamless interaction and knowledge exchange between distributed intelligent entities.

\section{Significance}

More than just a nominal definition:
In AI4S, "AI" denotes a data analysis methodology—specifically, using AI algorithms to address high-dimensional data challenges in scientific research.
In Agent4S, "Agent" represents a new productivity tool—Agent4S signifies agent-driven automation and intelligence in scientific research, emerging as a transformative productivity paradigm for data acquisition and processing.
The connection between AI4S and Agent4S in the entire scientific research process: Classical AI4S serves as the fundamental building blocks of algorithmic agents in the data processing stage within the Agent4S paradigm. Both fall under the category of applying AI to scientific research.
This analysis, from an AI technical perspective, provides concrete implementation and development frameworks for intelligent scientific research—moving beyond mere macro-level visions of the past. For AI practitioners, it clarifies how to advance AI technologies and integrate them with scientific workflows; for domain scientists, it identifies clear entry points for AI integration in their fields. Additionally, by outlining the scientific challenges at each developmental level of the Agent4S hierarchy, it guides researchers on aligning their work with AI advancements to systematically elevate the intelligence levels of research processes.

This work resolves the confusion between algorithms and intelligence: fundamentally, the technical essence of AlphaFold differs from that of cutting-edge intelligent laboratories. In terms of scope, the AI4S algorithms behind AlphaFold—at their core computational methods—represent a single component within the scientific research workflow and constitute a subset of the future Agent4S framework.
Regarding the classification of the Fifth Scientific Paradigm, numerous prior taxonomies existed but primarily focused on the degree of research automation, serving more as automation ratings rather than providing actionable technical roadmaps or trend analyses. Our classification innovatively integrates the technological evolution of Agents with the levels of research automation/intelligence, clearly delineating future development trajectories and critical challenges. This approach fundamentally advances the transformation toward the Fifth Scientific Paradigm by bridging theoretical visions with practical implementation frameworks.

In fact, the concept of the Fifth Scientific Paradigm was proposed early on \cite{li2024ai4r}, which outlined its characteristics:
\begin{enumerate}
    \item Full integration of artificial intelligence into scientific, technological, and engineering research, enabling knowledge automation and intelligence across the entire research process;
    \item Human-machine symbiosis, where emergent machine intelligence becomes an integral part of research, giving rise to tacit knowledge and machine-generated hypotheses;
    \item Focus on complex systems as primary research objects, effectively addressing combinatorial explosion problems with extremely high computational complexity;
    \item Orientation toward non-deterministic problems, where probabilistic and statistical reasoning play an expanded role in research;
    \item Cross-disciplinary collaboration as the mainstream research mode, achieving integration of the first four paradigms—especially the convergence of first-principles model-driven and data-driven approaches;
    \item Heavy reliance on large-scale platforms characterized by large models, with close integration between scientific research and engineering implementation.
\end{enumerate}

These characteristics are incisive, yet the paradigm lacked a strict definition at the time—due to the underdevelopment of agent technologies, many past AI algorithms for data processing were essentially refinements within the data-driven Fourth Scientific Paradigm but were erroneously categorized as part of the Fifth. From the Agent4S perspective, we can now naturally distinguish between "improvements within the data-driven paradigm" and "scientific paradigm shifts brought by new productivity tools."

In summary, both AI4S and Agent4S represent applications of AI in scientific research—where AI4S positions "AI" as a data analysis methodology, using AI algorithms to address high-dimensional data challenges and resolve the historical contradiction between data dimensionality and computational methods within the Fourth Scientific Paradigm. In contrast, Agent4S defines "Agent" as a new productivity tool, enabling agent-driven automation and intelligence in scientific research to tackle the contradiction between information richness and the productivity limitations of past paradigms.

It refers to using Agent technology to drive automation and intelligent scientific research, addressing the contradiction between information richness and productivity under past paradigms.

More than a nominal definition:
In AI4S, "AI" denotes a data analysis methodology—specifically, using AI algorithms to address high-dimensional data challenges in scientific research.
In Agent4S, "Agent" represents a new productivity tool—Agent4S signifies agent-driven automation and intelligence in research, emerging as a transformative paradigm for data acquisition and processing.

\printbibliography

\end{document}